\documentclass{article}

\usepackage{microtype}
\usepackage{graphicx}
\usepackage{subfigure}
\usepackage{booktabs} 

\usepackage{hyperref}
\graphicspath{{media/}}
\usepackage{csquotes}
\usepackage{enumitem}
\usepackage{float}
\usepackage{tabularx}
\usepackage{diagbox}
\usepackage{multicol}
\usepackage{amsmath}
\usepackage{multirow}
\usepackage{arydshln}
\usepackage{caption}
\usepackage{placeins}
\usepackage{subcaption}
\usepackage{fancyhdr}

\usepackage[accepted]{icml2024}

\usepackage{amsmath}
\usepackage{amssymb}
\usepackage{mathtools}
\usepackage{amsthm}
\usepackage{multirow}

\usepackage[capitalize,noabbrev]{cleveref}

\theoremstyle{plain}

\theoremstyle{definition}

\theoremstyle{remark}

\DeclareMathOperator*{\argmax}{arg\,max}

\usepackage[textsize=tiny]{todonotes}

\usepackage{authblk}

\title{Fine-Tuning or Retrieval? \\ Comparing Knowledge Injection in LLMs}

\author{Oded Ovadia \thanks{Corresponding author.} \thanks{Equal contribution.}}
\newcommand\CoAuthorMark{\footnotemark[\arabic{footnote}]}
\author{Menachem Brief\protect\CoAuthorMark}
\author{Moshik Mishaeli}
\author{Oren Elisha}

\affil{\{odedovadia,t-mbrief,mmishaeli,oren.elisha\}@microsoft.com \\ Microsoft, Israel}
\date{}

\begin{document}
\maketitle

\begin{abstract}
Large language models (LLMs) encapsulate a vast amount of factual information within their pre-trained weights, as evidenced by their ability to answer diverse questions across different domains. However, this knowledge is inherently limited, relying heavily on the characteristics of the training data. Consequently, using external datasets to incorporate new information or refine the capabilities of LLMs on previously seen information poses a significant challenge. In this study, we compare two common approaches: unsupervised fine-tuning and retrieval-augmented generation (RAG). We evaluate both approaches on a variety of knowledge-intensive tasks across different topics. Our findings reveal that while unsupervised fine-tuning offers some improvement, RAG consistently outperforms it, both for existing knowledge encountered during training and entirely new knowledge. Moreover, we find that LLMs struggle to learn new factual information through unsupervised fine-tuning, and that exposing them to numerous variations of the same fact during training could alleviate this problem.
\end{abstract}


\section{Introduction}
Large language models (LLMs) are able to capture vast amounts of factual information \cite{petroni2019language,cohen2023crawling,hu2023survey}. LLMs exhibit a remarkable level of knowledge in various domains due to their massive pre-training datasets. However, there are two significant limitations to this knowledge. First, it is static and does not update with time. Second, it is non-specific and thus may lack nuanced expertise in particular domains. While these are two different problems, they are deeply related since their solution is the same: enhancing the model's knowledge. 

Recently, the idea of adapting LLMs to particular domains and updating their knowledge has become increasingly common \cite{yu2022survey}. Various models have been suggested to improve factual knowledge and capabilities in diverse fields such as healthcare \cite{singhal2023large,singhal2023towards,wu2023pmc}, finance \cite{wu2023bloomberggpt,yang2023fingpt}, and law \cite{huang2023lawyer,nguyen2023brief}.

In this work, we focus on the evaluation of a model's knowledge and its ability to memorize, understand, and retrieve factual data. We aim to understand the concept of \textit{knowledge injection} \cite{wang2020k,chen2022knowprompt,liu2020k,lauscher2020common}. Given some knowledge base in the form of a text corpus, what is the best way to teach a pre-trained model this knowledge? 

One way to add knowledge to a pre-trained model is through fine-tuning. With fine-tuning, we continue the model's training process and adapt it using task-specific data. By exposing the model to a specific knowledge base, we expect the model weights to adapt accordingly. This process is meant to optimize the model for targeted applications, enhancing its performance and contextual relevance in specialized domains. 

Another method to enhance a model's knowledge base is through the use of in-context learning (ICL) \cite{chen2021meta,radford2019language,min2021metaicl,lampinen2022can}. The main idea behind ICL is to improve the performance of pre-trained LLMs on new tasks by modifying the input query to the model without directly changing the weights of the model. One form of ICL is retrieval augmented generation (RAG) \cite{lewis2020retrieval,Neelakantan2022TextAC}. RAG uses information retrieval techniques to enable LLMs to obtain relevant information from a knowledge source and incorporate it into generated text. 

This study aims to evaluate the knowledge injection capabilities of LLMs through a comparison of fine-tuning and RAG. To illustrate the rationale, let us use an analogy. Consider three college students taking a test on a specific topic. All had access to class materials but didn't know the topic beforehand. The first student had the textbook only during the test, the second had pre-test access and studied, and the third lost access upon the test announcement. Who would probably perform better?

\section{Background}\label{sec:motivation}

\begin{figure*}
  \includegraphics[width=0.98\textwidth]{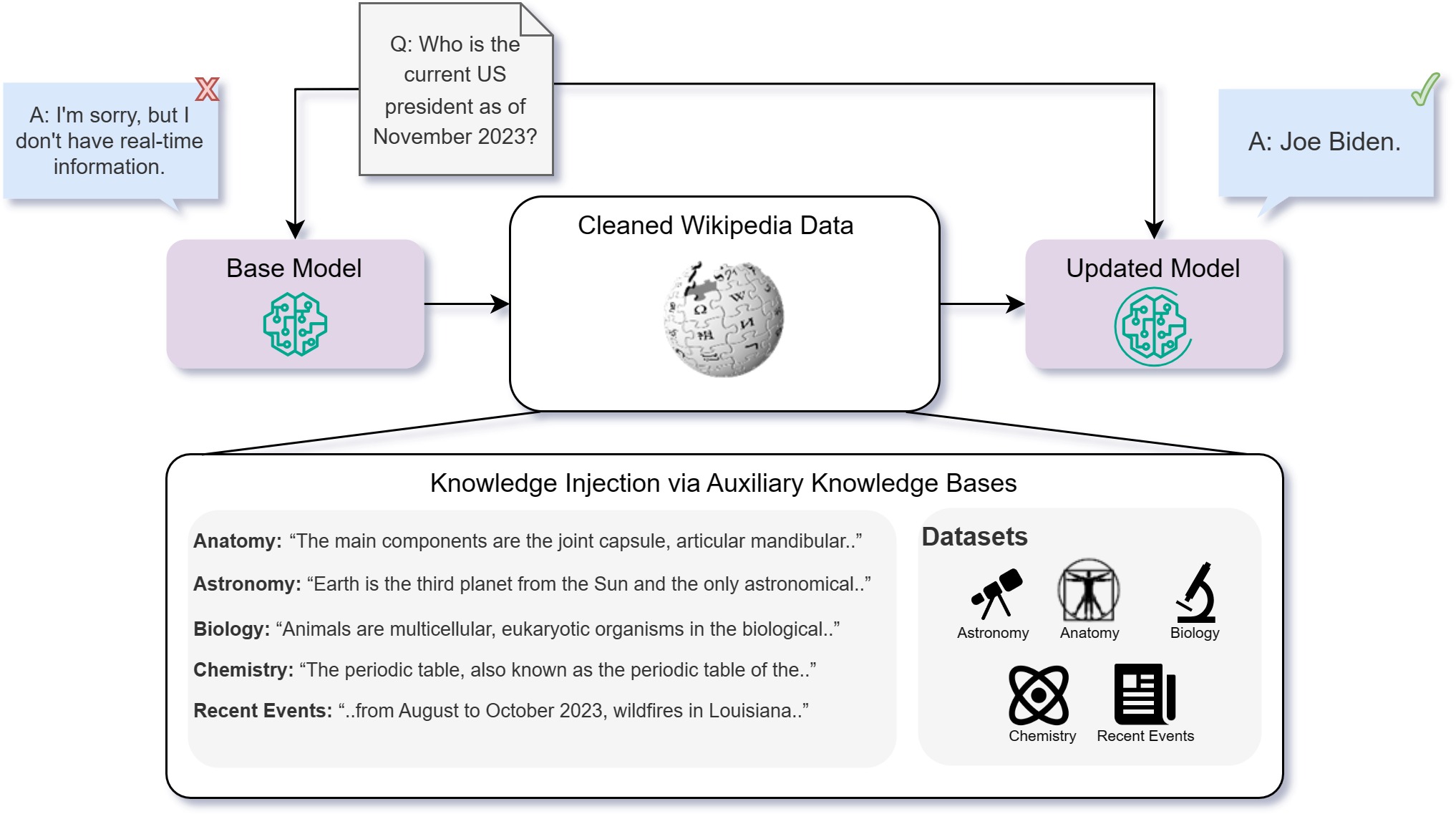}
  \caption{A visualization of the knowledge injection framework.}
\end{figure*}

To assess \textit{knowledge injection}, we must first understand what \textit{knowledge} means for LLMs.

\textbf{Knowledge and Language Models} \quad  Defining knowledge is a complex philosophical task far beyond the scope of this research. However, we can examine what factual knowledge means in the context of language models. If a model knows a fact, it can accurately and consistently answer questions about it. Furthermore, it can reliably distinguish between true and false statements related to this fact. We can then extend this definition to a whole knowledge base, not just a single fact. 

Mathematically, let $\mathcal{Q} = \{q_n\}_{n=1}^{N}$ be a set of $N$ multiple choice factual questions, where each question has $L$ possible answers and exactly one correct answer. Let $\mathcal{A} = \{(a_n^1, \ldots , a_n^L)\}_{n=1}^{N}$ be the corresponding set of possible answers, and $\mathcal{C} = \{c_n\}_{n=1}^{N}$ be the correct answers. 

Let $\mathcal{M}$ be a language model. We denote by $\mathcal{M}(q_n) \in \{a_n^1, \ldots , a_n^L\}$ the predicted answer of the model to the $n$-th question.

We define the \textit{knowledge score} $\mathcal{L}$ of $\mathcal{M}$ in relation to $\mathcal{Q}$ to be the standard accuracy score:
\begin{equation}\label{eq:knowledge_score}
    \mathcal{L}_{\mathcal{M}, \mathcal{Q}} : = \frac{\# \{q_n | \; \mathcal{M}(q_n) = c_n \}}{N}.
\end{equation}

We say that the model $\mathcal{M}$ possesses \textit{any} knowledge regarding the set of questions $\mathcal{Q}$ if the following holds:

\begin{equation}\label{eq:knowledge}
    \mathcal{L}_{\mathcal{M}, \mathcal{Q}} > \frac{1}{L}.
\end{equation}

\noindent In simpler terms, the model can consistently give correct answers, outperforming a simple random guessing baseline. Naturally, if the knowledge score $\mathcal{L}_{\mathcal{M}, \mathcal{Q}}$ is higher for one model compared to another, then we assert that the former is more knowledgeable with regards to $\mathcal{Q}$ compared to the latter.

\textbf{Previously Seen Knowledge} \quad One important distinction to make is between knowledge that the model has been exposed to before during pre-training as opposed to entirely new facts. Considering the size of modern LLM training sets, they cover a vast amount of information available through web-sourced text. As a result, even in niche domains, the goal of knowledge injection is not necessarily to teach the model entirely new facts but rather to "refresh" its memory by inducing a bias toward a particular domain.

\textbf{Knowledge and Reasoning} \quad We emphasize that this knowledge evaluation framework for LLMs is imperfect. Importantly, it doesn't address other quality metrics influencing a model's response. Creating a purely knowledge-intensive dataset without involving some level of reasoning is challenging. Consequently, a model with robust reasoning abilities might excel on unfamiliar knowledge-intensive tasks by making "educated guesses" in a multiple-choice exam. Therefore, any evaluation of knowledge in LLMs should consider this, with results seen as part of a broader range of benchmarks for reasoning \cite{sakaguchi2021winogrande}, reading comprehension \cite{dua2019drop}, and general language abilities \cite{srivastava2022beyond}. However, this evaluation framework still strongly emphasizes factual information above all else.

\textbf{Causes for Factual Errors} \quad There are many possible reasons for the failure of models to answer factual questions accurately. In \cite{wang2023survey}, Wang \textit{et al.} introduce a taxonomy of five main model-level causes:

\begin{itemize}[leftmargin=*]
    \item \textbf{Domain knowledge deficit}: A language model may lack comprehensive expertise in a specific domain to which it has not been exposed. For example, a model trained exclusively on texts written by William Shakespeare would perform poorly when asked about the works of Mark Twain.  
    
    \item \textbf{Outdated Information}: LLMs invariably have a cutoff date determined by their training dataset. Consequently, any events, discoveries, or changes occurring after the last training update will not be within the model's knowledge without access to external sources.
    
    \item \textbf{Immemorization}: Sometimes, a model is exposed to knowledge during its training process but does not retain it. This is especially true for rare facts that appear in the training dataset only scarcely \cite{kandpal2023large}. 
    
    \item \textbf{Forgetting}: Language models often undergo additional training after the pre-training phase (fine-tuning). In some cases, this might lead to a phenomenon called \textit{catastrophic forgetting} \cite{kirkpatrick2017overcoming,goodfellow2013empirical,chen2020recall,luo2023empirical}, where models lose some of the knowledge they had prior to the fine-tuning process. 
    
    \item \textbf{Reasoning Failure}: In certain instances, a language model might possess relevant knowledge about a fact but fail to utilize it properly. This is particularly evident in complex multi-step reasoning tasks \cite{tan2023can} or when posed with different questions about the same fact, resulting in disparate outcomes \cite{berglund2023reversal}.
\end{itemize}

We observe that most of these issues arise during the pre-training phase, with catastrophic forgetting being the notable exception. Hence, many LLMs will suffer from factual errors of this kind regardless of any post-training process.

\section{Injecting Knowledge to Language Models}

Following the background given in \cref{sec:motivation}, it is clear that general pre-training is insufficient for many knowledge-intensive tasks. To solve this, an additional post-processing step is essential to augment the knowledge of a pre-trained model. This step is often reffered to as \textit{knowledge injection} \cite{wang2020k,chen2022knowprompt,liu2020k,lauscher2020common}.

In this section, we examine two widely used frameworks for knowledge injection: fine-tuning (FT) and retrieval augmented generation (RAG). We begin by formulating the knowledge injection problem, aiming to explain both methods using consistent terminology.

\subsection{Problem formulation}
In \cref{eq:knowledge_score,eq:knowledge}, we presented a formulation for knowledge in language models through the lens of question-answering (Q\&A). We now extend this formulation to the problem of knowledge injection using the same terminology.

Given a set of factual questions, there exists some text corpus containing information that is relevant to these questions. The central assumption of knowledge injection is that given full access to this corpus, it could serve as an auxiliary knowledge base and improve the model's performance on this set of questions. 

Mathematically, let $\mathcal{M}$ be a pre-trained model, and let  $\mathcal{Q}$ be a set of factual questions, as before. Now, assume we have a relevant auxiliary knowledge base $\mathcal{B}_\mathcal{Q}$. Our objective is to discover a transformation, denoted as $\mathcal{F}$, that, when applied, would enhance the knowledge about $\mathcal{Q}$:

\begin{equation}
    \mathcal{M'} := \mathcal{F}(\mathcal{M},\mathcal{B}_\mathcal{Q}) \quad s.t. \quad \mathcal{L}_{\mathcal{M'}, \mathcal{Q}} > \mathcal{L}_{\mathcal{M}, \mathcal{Q}}.
\end{equation}

In this work, we aim to compare two choices for $\mathcal{F}$: fine-tuning and RAG to see which option performs better in this problem. 


\subsection{Fine-Tuning}\label{subsec:fine_tuning}
Fine-tuning is the process of adjusting a pre-trained model on a specific, often narrower, dataset or task to enhance its performance in that particular domain. Here, it is vital to distinguish between different types of fine-tuning. FT techniques are commonly classified into supervised, unsupervised, and reinforcement learning (RL) based methods. We proceed by briefly reviewing these methods and their relation to the problem of knowledge injection.

\textbf{Supervised Fine-Tuning} \quad Supervised fine-tuning (SFT) requires sets of labeled input-output pairs. One of the most common SFT methods is instruction tuning \cite{wang2022super,mishra2021cross,ouyang2022training,taori2023alpaca}, which has emerged as one of the most powerful methods to improve model performance. With instruction tuning, the input is a natural language task description, and the output is an example of the desired behavior. Many current state-of-the-art LLMs have gone through instruction tuning after their pre-training phase. 

Instruction tuning has been shown to be very effective at improving the overall quality of the model, with a particular emphasis on its zero-shot and reasoning capabilities. However, despite these advantages, instruction tuning does not necessarily teach the model new knowledge \cite{ouyang2022training,chung2022scaling,mitra2023orca,chia2023instructeval,zhou2023lima}. As such, instruction tuning alone is not a viable solution to the knowledge injection problem.

\textbf{Reinforcemnt Learning} \quad Another form of FT relies on RL or RL-inspired optimization strategies to better align the model after its pre-training phase. A few prominent examples are reinforcement learning from human feedback (RLHF) \cite{OpenAI2023GPT4TR,touvron2023llama}, direct preference optimization (DPO) \cite{rafailov2023direct}, and proximal policy optimization (PPO) \cite{schulman2017proximal,tunstall2023zephyr}.

These techniques have been shown to be very useful, especially when used in conjunction with instruction tuning. However, similarly to instruction tuning, these methods focus on the overall quality of the response and its expected behavior and not necessarily on its breadth of knowledge. 

\textbf{Unsupervised Fine-Tuning} \quad The final FT strategy we discuss is unsupervised, meaning there are no available labels for the model to learn from. One common unsupervised FT technique is often referred to as \textit{continual pre-training} or \textit{unstructured} FT. 

In this method, the FT process is viewed as a direct continuation of the pre-training phase. We start with a saved checkpoint of the original LLM and train it in a causal auto-regressive manner, i.e., predicting the next token. One major difference in comparison to actual pre-training is the learning rate. Usually, one would need a much lower learning rate when continuing the pre-training of the model to avoid catastrophic forgetting \cite{kirkpatrick2017overcoming}. 

It is well known that LLMs store vast amounts of knowledge during their pre-training phase \cite{zhou2023lima}. So, it makes sense to continue this process in order to inject knowledge into the model. Hence, we use the unsupervised FT approach throughout this work and evaluate its efficacy in enhancing the model's capacity for learning new information.

\subsection{Retrieval Augmented Generation}
Retrieval augmented generation (RAG)~\cite{lewis2020retrieval} is a technique that expands LLMs' capabilities, especially in knowledge-intensive tasks, by using external knowledge sources. While the original formulation involved additional training per task, it has since been demonstrated~\cite{Neelakantan2022TextAC} that a pre-trained \textit{embedding} model can achieve improved performance with no additional training involved. 

The idea is that given an auxiliary knowledge base and an input query, we use the RAG architecture to find documents within the knowledge base that resemble the input query. These documents are then added to the input query, thus giving the model further context about the subject of the query. 

In practice, implementing the suggested architecture is quite straightforward: Given an auxiliary knowledge base $\mathcal{B}_\mathcal{Q}$ and a pre-trained embedding model $\mathcal{M}_e$, we create a dense vector representation (embedding) per document $b\in\mathcal{B}_\mathcal{Q}$ and store these in a vector store. Upon receiving a new query $q$, we use its embedding, $\mathcal{M}_e(q)$, to retrieve $q$'s top-$K$ closest neighbors, $\mathbf{b}_q = \{b_k\}_1^K$, according to dot-product ranking. We then update $q$ to be $\tilde{q} = \mathbf{b}_q\Vert q$, where $\Vert$ denotes string concatenation. Finally, we return $\mathcal{M}(\tilde{q})$ as the model's output.

\section{Knowledge Base Creation}

\begin{table*}[!htb]
\caption{Results for the MMLU datasets described in \cref{subsec:tasks} in terms of log-likelihood accuracy (\cref{eq:log_likelihood}). }
\vskip 0.15in
\begin{center}
\begin{small}
    \begin{tabular}{llcccc}
  \toprule
    Task & Model & Base model & Base model + RAG & Fine-tuned & Fine-tuned + RAG \\ 
    \midrule
    
    \multirow{3}{*}{Anatomy (0-shot)} & Mistral 7B & 0.556 & \textbf{0.681} & 0.570 & 0.659 \\
    & Llama2 7B & 0.393 & \textbf{0.489} & 0.430 & \textbf{0.489} \\
    & Orca2 7B  & 0.607 & \textbf{0.637} & 0.600 & \textbf{0.637} \\
    & & & & & \\
    \multirow{3}{*}{Anatomy (5-shot)} & Mistral 7B & 0.600 & \textbf{0.681} & 0.622 & 0.674 \\
    & Llama2 7B  & 0.467 & \textbf{0.563} & 0.496 & 0.548 \\
    & Orca2 7B  & 0.570 & 0.659 & 0.593 & \textbf{0.674} \\
    \midrule
    
    \multirow{3}{*}{Astronomy (0-shot)} & Mistral 7B  & 0.625 & 0.678 & 0.651 & \textbf{0.697} \\
    & Llama2 7B & 0.401 & 0.467 & 0.487 & \textbf{0.520} \\
    & Orca2 7B & 0.645 & \textbf{0.750} & 0.651 & \textbf{0.750} \\
    & & & & & \\
    \multirow{3}{*}{Astronomy (5-shot)} & Mistral 7B & 0.658 & \textbf{0.724} & 0.651 & 0.697 \\
    & Llama2 7B &  0.401 & 0.474 & 0.447 & \textbf{0.520} \\
    & Orca2 7B &  0.664 & \textbf{0.763} & 0.664 & 0.743 \\
    \midrule
    
    \multirow{3}{*}{College biology (0-shot)} & Mistral 7B & 0.681 & 0.757 & 0.701 & \textbf{0.764}  \\
    & Llama2 7B & 0.438 & \textbf{0.493} & 0.458 & 0.465  \\
    & Orca2 7B  & 0.583 & \textbf{0.639} & 0.604 & 0.632  \\
    & & & & & \\
    \multirow{3}{*}{College biology (5-shot)} & Mistral 7B & 0.722 & \textbf{0.778} & 0.736 & 0.771  \\
    & Llama2 7B  & 0.451 & \textbf{0.521} & 0.424 & 0.479  \\
    & Orca2 7B  & 0.604 & \textbf{0.660} & 0.625 & 0.653   \\ 
    \midrule

    \multirow{3}{*}{College chemistry (0-shot)} & Mistral 7B & 0.470 & \textbf{0.500} & 0.490 & \textbf{0.500} \\
    & Llama2 7B & 0.310 & 0.380 & 0.390 & \textbf{0.390}  \\
    & Orca2 7B  & 0.370 & \textbf{0.440} & 0.370 & 0.390  \\
    & & & & & \\
    \multirow{3}{*}{College chemistry (5-shot)} & Mistral 7B &  0.470 & \textbf{0.540} & 0.500 & 0.500  \\
    & Llama2 7B  & 0.370 & 0.380 & 0.360 & \textbf{0.390}  \\
    & Orca2 7B   &  0.430 & \textbf{0.470} & 0.370 & 0.380 \\ 

    \midrule
    \multirow{3}{*}{Prehistory (0-shot)} & Mistral 7B & 0.713 & \textbf{0.750} & 0.719 & 0.731   \\
    & Llama2 7B & 0.448 & \textbf{0.481} & 0.457 & 0.478 \\
    & Orca2 7B  &  0.642 & \textbf{0.679} & 0.673 & 0.673 \\
    & & & & & \\
    \multirow{3}{*}{Prehistory (5-shot)} & Mistral 7B & 0.722 & \textbf{0.762} & 0.725 & \textbf{0.762}   \\
    & Llama2 7B  & 0.515 & 0.531 & 0.503 & \textbf{0.537}  \\
    & Orca2 7B   &  0.664 & \textbf{0.698} & 0.667 & 0.694  \\ 
    \bottomrule
  \end{tabular}
  \end{small}
  \end{center}
  \label{tab:mmlu}
\end{table*}

\begin{table*}[!htb]
    \caption{Current events results. Models that were fine-tuned on the original dataset are labeled as \textit{FT-reg}, while those trained on the dataset with multiple paraphrases are labeled as \textit{FT-par}.}
    \vskip 0.15in
    \begin{center}
    \begin{small}
        \begin{tabular}{lcccccc}
        \toprule
         & Base model & Base model + RAG & FT-reg & FT-par & FT-reg + RAG & FT-par + RAG \\
        \midrule
        Mistral 7B & 0.481 & \textbf{0.875} & 0.504 & 0.588 & 0.810 & 0.830 \\
        Llama2 7B  & 0.353 & \textbf{0.585} & 0.219 & 0.392 & 0.326 & 0.520 \\
        Orca2 7B & 0.456 & \textbf{0.876} & 0.511 & 0.566 & 0.820 & 0.826 \\
        \bottomrule
    \end{tabular}
    \end{small}
    \end{center}
    \label{tab:current_events}
\end{table*}

\subsection{Task Selection and Rationale}\label{subsec:tasks}

\textbf{MMLU Benchmark} \quad
To properly evaluate the capabilities of LLMs on knowledge-intensive tasks, we selected four distinct tasks from the Massively Multilingual Language Understanding Evaluation (MMLU) benchmark~\cite{hendryckstest2021} in the topics of anatomy, astronomy, college biology, college chemistry and prehistory. The chosen tasks were selected based on their emphasis on factual knowledge and the minimal reliance on reasoning. As a heuristic, we opted for tasks where the questions are short and involve no context. In practice we selected four STEM subjects as well as one humanities subject, to ensure the evaluation is not limited to certain fields. Note that prehistory involves questions spanning all non-modern history. This approach aims to enable us to test LLM proficiency in comprehending and manipulating information in isolation from its reasoning processes. 

\textbf{Current Events Task} \quad
To further isolate LLMs' abilities to learn new knowledge, we created a task comprising multiple-choice questions about current events. This task includes multiple-choice questions about events that occurred after the cutoff of the various models' training data. Specifically, we focused on "current events" from the USA, in the time span of August-November 2023, that are included in the relevant Wikipedia indexes\footnote{\url{https://en.wikipedia.org/wiki/Category:2023_events_in_the_United_States_by_month}}. This method enables us to mostly guarantee that the models have not been exposed to these facts, thus allowing us to directly test knowledge injection capabilities.   

\subsection{Data Collection and Preprocessing}
\label{subsec:Data collection and preprocessing}

To effectively evaluate the LLMs' performance on these knowledge-intensive tasks, a comprehensive auxiliary dataset was collected by scraping relevant articles per topic from Wikipedia. The rationale behind selecting Wikipedia as the primary source of knowledge is its broad coverage of relevant topics and its reliability as a repository of crowd-verified knowledge. All articles pertinent to the tasks were retrieved via the official Wikipedia API\footnote{\url{https://www.mediawiki.org/wiki/API:Main_page}} by identifying the relevant central page per topic.

Subsequently, a rigorous cleaning process was utilized to transform the data from raw subsections to clean chunks. This step was done with the "wikiextractor" tool~\cite{Wikiextractor2015}. The division into small, clean (e.g., remove HTML, URLs, etc.) chunks was aimed at enhancing the evaluation of the LLMs' understanding across various knowledge domains and aiding the LLMs in the fine-tuning process.

\subsection{Current Events Task Creation}
\label{subsec: custom dataset curation}
After collecting the relevant chunks from Wikipedia, we created a new multiple-choice dataset with the help of GPT-4~\cite{OpenAI2023GPT4TR}. First, we removed any small chunks. For each remaining chunk in the corpus, GPT-4 was instructed to create four highly specific, high-quality multiple-choice questions with only one correct answer. By specific, we mean that the question can be answered without knowledge of which context the question refers to and with minimal ambiguity. Next, GPT-4 was asked to select the two most specific of the four. This was followed by a manual evaluation and verification step. In total, this resulted in 910 new questions.  

\subsection{Paraphrases Generation}
After creating the dataset, we utilized GPT-4 to generate augmentations of the dataset. We instructed GPT-4 to provide paraphrased versions of the input data that fully retain the information while being reworded. Each paraphrasing iteration was done with a different seed to ensure variety.

We selected 240 chunks at random for each task and created two paraphrases per chunk. These were set aside to be used as validation sets for hyperparameter tuning. For the current events dataset, we created ten paraphrases for each chunk used in the fine-tuning process described in~\cref{sec:paraphrase}. 

\section{Experiments and Results}\label{subsec: experimental setup}

\textbf{Experimental Framework} \quad We used the popular LM-Evaluation-Harness~\cite{eval-harness} repository to evaluate the performance of LLMs on the selected knowledge-intensive tasks. LM-Evaluation-Harness is a robust benchmarking tool that currently serves as the industry standard for model evaluation and is the basis of the HuggingFace leaderboard\footnote{\url{https://huggingface.co/spaces/HuggingFaceH4/open_llm_leaderboard}}. Leveraging this platform ensured a standardized evaluation framework and allowed consistent comparison across models, methods, and datasets. More importantly, by using the industry standard for evaluation, we could avoid any differences stemming from prompt engineering and formatting issues and replicate the reported baseline results for each model. 

\textbf{Model Selection} \quad
We chose three models for inference evaluation: Llama2-7B \cite{touvron2023llama}, Mistral-7B \cite{jiang2023mistral}, and Orca2-7B \cite{mitra2023orca}.
The choice of these models was meant to represent the most popular open-source base models and an instruction-tuned model across various baseline capabilities. 
Additionally, we selected \textit{bge-large-en}~\cite{bge_embedding} as the embedding model for the RAG component and used FAISS~\cite{johnson2019billion} as its vector-store. This embedding model is currently the SOTA of open-source embedding models, according to the HuggingFace MTEB leaderboard\footnote{\url
{https://huggingface.co/spaces/mteb/leaderboard}}.  

\begin{figure}[tb]
\begin{center}
\centerline{\includegraphics[width=\columnwidth]{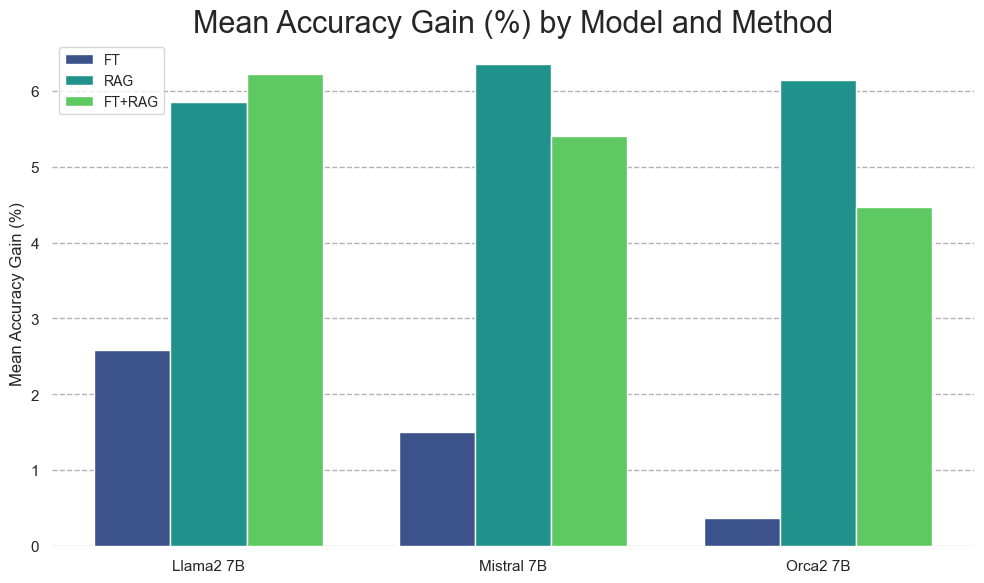}}
\caption{The relative accuracy gain (as explained in~\cref{eq:relative acc}) for each knowledge-injection method, averaged (columnwise) across all experiments in~\cref{tab:mmlu}.}
\label{fig:barplot}
\end{center}
\vskip -0.25in
\end{figure}

\textbf{Configuration Variations} \quad
Our evaluation included multiple configurations, with a grid-search over them, to allow for more comprehensive benchmarking. \\
Firstly, we compared the baseline and fine-tuned models and their performance with the RAG component. Secondly, we explored the optimal number of text chunks to add to the context in RAG. Specifically, different values of $K\in\{0,\ldots, 5\}$ were employed to analyze the impact on model performance. Finally, we explored 5-shot performance vs. 0-shot. 

\textbf{Training Setup} \quad  We trained all of the models using the unsupervised training procedure described in \cref{subsec:fine_tuning}. For each dataset, we 
divided the auxiliary knowledge base into equal chunks of size $256$ by concatenating or splitting the original chunks based on their length. We also added two special tokens, $<$BOS$>$ and $<$EOS$>$, to demarcate the original chunks' beginnings and ends to preserve the documents' structure. 

The models were trained using learning rates between $1\times {10}^{-6}$ and $5\times {10}^{-5}$, which were found through a hyperparameter search. All models were trained on 4 NVIDIA A-100 GPUs for a maximum of 5 epochs and a batch size of 64.

\textbf{Evaluation method} \quad All evaluations were done by appending each of the multiple-choice options to the question, followed by passing the concatenation through the model to get a log probability score per option. The highest score was interpreted as the model's choice and used for accuracy calculation. More formally, this means that in~\cref{eq:knowledge_score} we say that $\mathcal{M}(q_n) = c_n$ if:
\begin{equation}\label{eq:log_likelihood}
   c_n = \argmax_l \{\mathcal{M}(q_n \Vert a^1_n), \ldots, \mathcal{M}(q_n \Vert a^L_n)\},
\end{equation} 
where $\mathcal{M}(q_n \Vert a^l_n) = \log P_{\mathcal{M}}(q_n \Vert a^l_n)$.

\textbf{MMLU Results} \quad For each task and model, we compared four approaches: using just the base model, RAG, FT, and finally combining FT and RAG by using the fine-tuned model as the generator. Furthermore, we tested the MMLU tasks using both 0-shot and 5-shot scenarios. The full results are shown in ~\cref{tab:mmlu}. An aggregation of the relative accuracy gain, i.e., 
\begin{equation}
\label{eq:relative acc}
    (\mathcal{L}_{\mathcal{M'}, \mathcal{Q}} - \mathcal{L}_{\mathcal{M}, \mathcal{Q}})/{\mathcal{L}_{\mathcal{M}, \mathcal{Q}}},
\end{equation} where $\mathcal{M}$ is the base model and $\mathcal{M'}$ is the  knowledge-injected model, is shown in ~\cref{fig:barplot}.

In all cases, RAG performed significantly better compared to the base models. Furthermore, using RAG with the base model as the generator was consistently better than only fine-tuning. In some cases, using the fine-tuned model instead of the base model as the generator in the RAG pipeline improved results even further. However, this is not consistent and thus demonstrates the inherent instability of fine-tuning. Additionally, we found that the 5-shot approach boosts the results by a small margin in most cases, with a similar trend being observed in all of the different approaches. 

\textbf{Current Events Results} \quad The evaluation on the current events task is shown in~\cref{tab:current_events}. RAG proves particularly effective due to the one-to-one correspondence between the questions and the auxiliary dataset (see~\cref{subsec: custom dataset curation}). Fine-tuning is not competitive with RAG. However, fine-tuning with multiple paraphrases still provides a significant improvement over the baseline. We note that combining RAG with fine-tuning shows inferior performance compared to RAG alone.

It is worth noting that although the questions are based on information the models were not exposed to during training,  the results of the base models surpass $\frac{1}{L} = 0.25$. This can partially be explained by the models using reasoning and/or pre-existing knowledge when answering questions that are not independent of the past information. Some examples of this can be found in~\cref{appendix:current_events_analysis}.

\textbf{Fine-Tuning vs. RAG:} In the results of both the MMLU and current events tasks, a significant advantage for RAG over fine-tuning is evident. While fine-tuning improved results compared to the base model in most cases, it was not competitive with the RAG approach. 

Several factors might contribute to this behavior. Firstly, RAG not only adds knowledge to a model but also incorporates context relevant to the question, a feature lacking in fine-tuning. Additionally, fine-tuning may impact other capabilities of the model due to a degree of catastrophic forgetting. Finally, it's plausible that unsupervised fine-tuned models might benefit from further alignment through supervised or RL-based fine-tuning, as evidenced by the vastly improved performance of Orca2 over the base Llama2.  

\section{The Importance of Repetition}
\label{sec:paraphrase}

Unlike the other tasks, where the model has been exposed to aspects related to the topic during pretraining, \textit{current events} includes new information. In this case, standard regular fine-tuning not only did not improve the performance of Llama2 but also significantly degraded it. To improve the fine-tuning results, we explored augmentation of the data using paraphrases. 

\begin{figure}[!htb]
\begin{center}
\centerline{\includegraphics[width=\columnwidth]{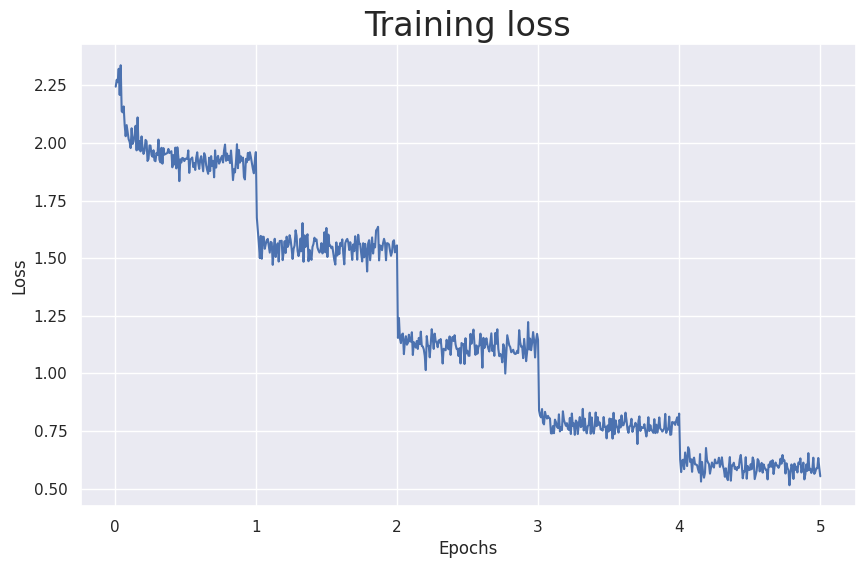}}
\caption{Training loss over time for Mistral-7B.}
\label{fig:loss}
\end{center}
\vskip -0.2in
\end{figure}

\begin{figure}[!ht]
  \includegraphics[width=\columnwidth]{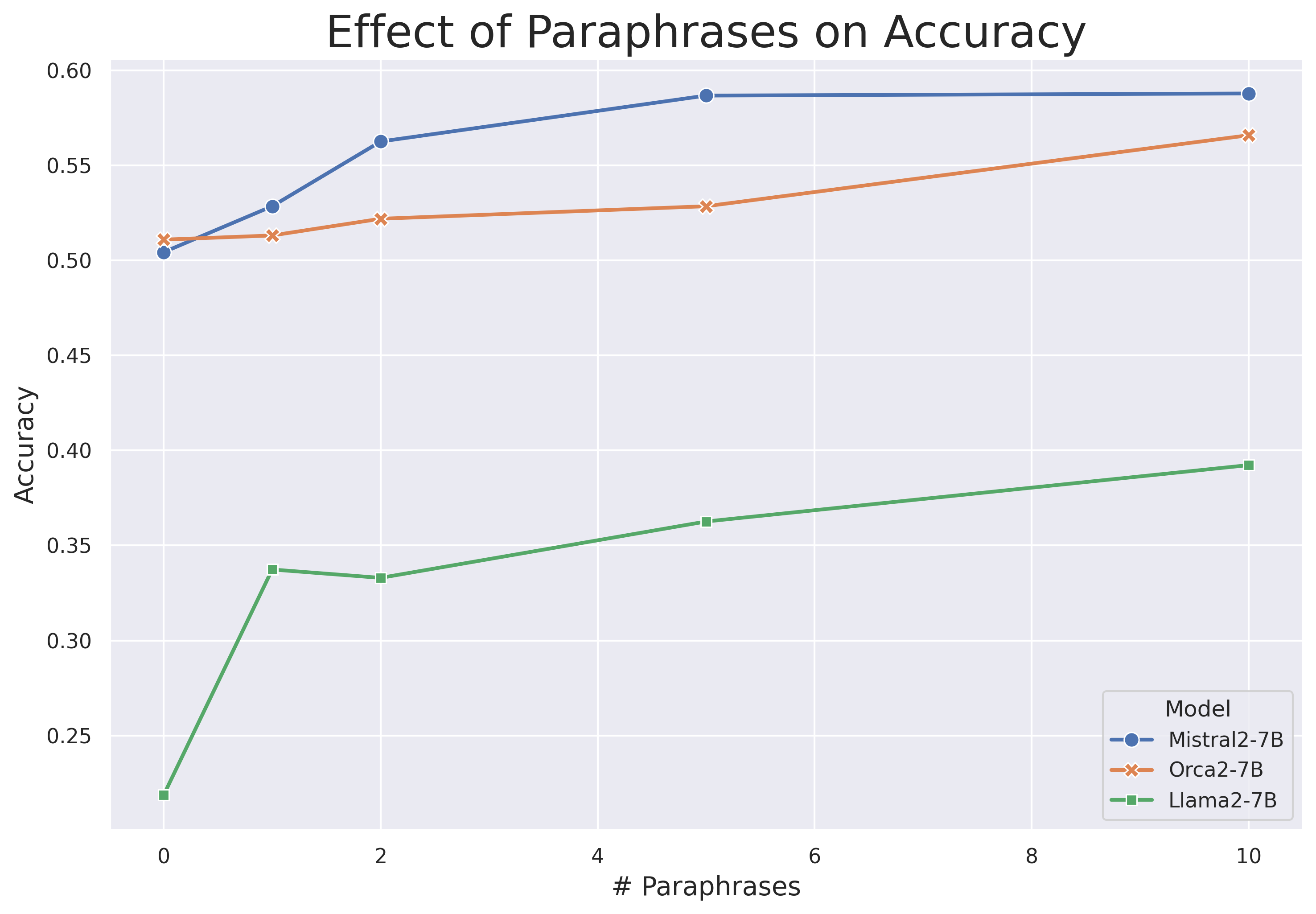}
  \caption{Model accuracy on the \textit{current events} task as a function of the number of paraphrases.}
  \label{fig: effect of paraphrases}
\end{figure}

\textbf{Data Augmentation} \indent
Data augmentation is a well-established method for enhancing the performance of language models and has been surveyed extensively~\cite{Shorten2021TextDA}. Using generative models for augmentations has also been used successfully to improve classification models in the past~\cite{sharma2022systematic}.
An example of data augmentation using paraphrasing can be found in ~\cref{appendix:parphrases}.

\textbf{Monotonic Improvement} \indent
This approach resulted in notable improvements in our results, showcasing a direct correlation between the number of paraphrases utilized and the models' accuracy.
Our experimentation revealed a compelling trend, shown in~\cref{fig: effect of paraphrases}. For all models tested, the accuracy was a monotonically increasing function of the number of paraphrases used. This observation strongly suggests the positive impact of paraphrase augmentation, yielding information repetition, on the model's ability to comprehend and generalize new knowledge from limited data.

\textbf{Learning New Information} \indent
In~\cref{fig:loss}, we can see an interesting phenomenon observed throughout our experiments. After each epoch, i.e., completing another iteration over the entire dataset, the training loss drops significantly. This is consistent with what is known about LLMs memorizing the data during training and overfitting~\cite{Tirumala2022MemorizationWO}. 

Our hypothesis is as follows:
\begin{quote}
    In order to teach pre-trained LLMs \textbf{new} knowledge, the knowledge must be repeated in numerous ways.
\end{quote}

\noindent This is well known for LLM pre-training \cite{kandpal2023large}, and we see in this case that this holds for fine-tuning as well. The rationale for this hypothesis is that mere memorization of sentences does not entail knowledge of their content, as was already shown in~\cite{berglund2023reversal}. By providing the information in numerous forms (like the data augmentation process we used), the various relationships in the data (e.g., $a\implies b,\: b\rlap{$\quad\not$}\implies c$) stand a higher chance of appearing naturally. We believe this can potentially both increase $\mathcal{L}_{\mathcal{M}, \mathcal{Q}}$ in general, as well as ameliorate Berglund et al.'s \textit{Reversal Curse}. While promising, this result still warrants further research.

\section{Conclusion and Future Work}
Large language models possess vast amounts of knowledge on various topics. In this work, we tested their capability to adapt to new knowledge: both specialized and completely unseen. This is among the first studies to compare two prominent approaches in this domain, namely fine-tuning and retrieval augmented generation. While fine-tuning can be useful for many use-cases, we found that RAG is a more reliable choice for knowledge injection.

Some aspects of this work still warrant further research. For example, we focused on unsupervised training as our primary fine-tuning method, as opposed to instruction-tuning or RL-based methods. Researching combinations of various techniques, with diverse auxiliary knowledge bases, may yield improved results. This approach, combined with our hypothesis from~\cref{sec:paraphrase}, could further enhance our understanding of knowledge injection via FT.

While we believe that this work further enhances our understanding of knowledge in LLMs, there is a lot more work to be done in this field. Specifically, more research is required regarding the question of knowledge representation in LLMs, especially from a theoretical perspective. 

Finally, further efforts are needed to measure knowledge in LLMs. While we employed an empirical approach as described in~\cref{eq:knowledge}, it is important to explore other definitions and perspectives on knowledge as well, and extend upon this work.

\section{Limitations}
As in all machine learning applications, the choice of hyperparameters significantly impacts the results. We therefore strongly recommend optimizing all relevant hyperparameters for specific cases. 

\noindent We have supported our claims by running the experiments on three different models. However, generalization to other LLMs should be tested thoroughly. For example, GPT-4 achieves near perfect accuracy for some MMLU tasks~\cite{Nori2023CapabilitiesOG}, and thus further improvement is not applicable. 

Finally, while we chose various topics for the knowledge bases, all of our sources came from Wikipedia. Other datasets may yield different results, and must be evaluated carefully. 


\FloatBarrier
\bibliography{references}
\bibliographystyle{icml2024}

\newpage
\appendix
\onecolumn
\section{RAG Ablation Study}
As mentioned in \cref{subsec: experimental setup}, we compared various values of $K\in\{0,\ldots, 5\}$, shown in \cref{tab:rag_ablation}.We were unable to find an optimal value of $K$ per model, per $0/5$-shot, or per task. In fact, other than Anatomy that worked well with $K=2$ consistently, there seems to be no patterns that aid in predicting the performance per $K$, unlike the results presented in~\cite{lewis2020retrieval} for other setups. Moreover, the gap between the best and worst performing $K$s can be large.\\
\noindent Unfortunately, we must conclude that this additional hyperparameter is unstable. This is a downside of using RAG in practice, and the choice of $K$ cannot be ignored.    

\begin{table*}[!htb]
  \centering
\begin{tabular}{ll|ccccc}
\multirow{2}{*}{Task} & \multirow{2}{*}{Model} &  \multicolumn{5}{c}{\# Retrieved documents ($k$)} \\
 &  & 1 & 2 & 3 & 4 &5 \\
\midrule
\multirow{3}{*}{Anatomy (0-shot)} & Mistral 7B & 0.615 & \textbf{0.681} & 0.630 & 0.644 & 0.622 \\
 & Llama2 7B & 0.444 & \textbf{0.489} & 0.467 & 0.474 & 0.481 \\
 & Orca2 7B & 0.607 & \textbf{0.637} & 0.600 & 0.585 & \textbf{0.637} \\
 &  &  &  &  &  & \\
\multirow{3}{*}{Anatomy (5-shot)} & Mistral 7B & 0.659 & 0.667 & 0.659 & \textbf{0.681} & 0.674 \\
 & Llama2 7B & 0.496 & \textbf{0.563} & 0.541 & 0.526 & 0.526  \\
 & Orca2 7B & 0.630 & \textbf{0.659} & 0.600 & 0.600 & 0.600 \\
\midrule

\multirow{3}{*}{Astronomy (0-shot)} & Mistral 7B & 0.651 & \textbf{0.678} & \textbf{0.678} & 0.664 & 0.664 \\
& Llama2 7B & 0.447 & 0.434 & 0.447 & 0.434 & \textbf{0.467} \\
& Orca2 7B & 0.711 & 0.730 & 0.730 & \textbf{0.750} & 0.730 \\
 &  &  &  &  &  & \\
\multirow{3}{*}{Astronomy (5-shot)} & Mistral 7B & 0.704 & 0.684 & 0.658 & 0.684 & \textbf{0.724} \\
 & Llama2 7B & 0.461 & 0.447 & \textbf{0.474} & 0.428 & 0.454 \\
 & Orca2 7B & 0.730 & 0.737 & 0.750 & 0.743 & \textbf{0.763} \\
\midrule

\multirow{3}{*}{Biology (0-shot)} & Mistral 7B & 0.736 & 0.722 & \textbf{0.757} & 0.743 & 0.736 \\
& Llama2 7B & 0.438 & 0.472 & \textbf{0.493} & 0.479 & 0.472 \\
 & Orca2 7B & \textbf{0.639} & 0.618 & \textbf{0.639} & 0.625 & \textbf{0.639} \\
 &  &  &  &  &  & \\
\multirow{3}{*}{Biology (5-shot)} & Mistral 7B & 0.722 & \textbf{0.778} & \textbf{0.778} & 0.771 & 0.743 \\
& Llama2 7B & 0.500 & \textbf{0.521} & 0.507 & 0.465 & 0.472 \\
& Orca2 7B & 0.625 & 0.639 & 0.625 & \textbf{0.660} & \textbf{0.660} \\
\midrule

\multirow{3}{*}{Chemistry (0-shot)} & Mistral 7B & 0.450 & 0.470 & 0.470 & \textbf{0.500} & 0.470 \\
& Llama2 7B & 0.320 & 0.320 & 0.300 & \textbf{0.380} & 0.360 \\
& Orca2 7B & 0.370 & 0.420 & 0.400 & 0.410 & \textbf{0.440} \\
 &  &  &  &  &  & \\
\multirow{3}{*}{Chemistry (5-shot)} & Mistral 7B & \textbf{0.540} & 0.490 & 0.500 & 0.510 & 0.470 \\
& Llama2 7B & 0.280 & 0.320 & 0.340 & 0.340 & \textbf{0.380} \\
& Orca2 7B & 0.390 & 0.430 & 0.400 & 0.430 & \textbf{0.470} \\

\midrule

\multirow{3}{*}{Prehistory (0-shot)} & Mistral 7B & 0.728 & 0.725 & \textbf{0.750} & 0.735 & 0.728 \\
& Llama2 7B & \textbf{0.481} & 0.460 & 0.457 & 0.457 & 0.429 \\
& Orca2 7B & 0.648 & 0.645 & 0.660 & 0.670 & \textbf{0.679} \\
 &  &  &  &  &  & \\
\multirow{3}{*}{Prehistory (5-shot)} & Mistral 7B & 0.710 & 0.750 & 0.759 & 0.756 & \textbf{0.762} \\
& Llama2 7B & 0.512 & 0.485 & 0.525 & 0.519 & \textbf{0.531} \\
& Orca2 7B & 0.660 & 0.688 & 0.685 & \textbf{0.698} & 0.688 \\
\bottomrule
\end{tabular}
  \caption{RAG ablation study.}
  \label{tab:rag_ablation}
\end{table*}

\section{Paraphrase Examples}\label{appendix:parphrases}
Below is the prompt we used to generate paraphrases with GPT-4:

\begin{quote}
Your task is to paraphrase a text paragraph. The paragraph is given below. 

Make sure to keep the same meaning but change the wording. Do not change any factual information. 

Try to keep roughly the same length of the original text. 

Give NUM \_ PARAPHRASES different paraphrases for each text. 

These paraphrases should be as different from each other as possible.

Return a JSON formatted string with one key, called 'paraphrases', and a list of paraphrases.

Input paragraph: 
\end{quote}

Below are some examples. First, an example taken from the following Wikipedia page: "51st International Emmy Awards".

\begin{quote}
    \textbf{\underline{Paraphrase I}}

    The fifty-first edition of the International Emmy Awards Competition accepted submissions across all categories from December 7, 2022, until the closing date of February 16, 2023.

    \textbf{\underline{Paraphrase II}}
    
    The International Emmy Awards Competition, now in its 51st year, accepted submissions across all categories starting from December 7, 2022, until the deadline on February 16, 2023.

    \textbf{\underline{Paraphrase III}}

    The fifty-first edition of the International Emmy Awards Competition accepted submissions across all categories from December 7, 2022, until the closing date of February 16, 2023.

\end{quote}

The next example was taken from the following Wikipedia page: "2023 Indianapolis mayoral election".

\begin{quote}
    \textbf{\underline{Paraphrase I}}
    The mayoral election in Indianapolis took place on November 7, 2023, with preliminary elections occurring on May 2. The sitting mayor, Democrat Joe Hogsett, successfully ran for a third term. Both Hogsett and his Republican opponent, Jefferson Shreve, moved on to the main election.

    \textbf{\underline{Paraphrase II}}
    On November 7, 2023, citizens of Indianapolis cast their votes to elect their Mayor, following primary elections on May 2. Joe Hogsett, the Democrat already in office, won his bid for a third term. Hogsett and the Republican candidate, Jefferson Shreve, were the two contenders in the final electoral round.

    \textbf{\underline{Paraphrase III}}

    The mayoral election in Indianapolis took place on the 7th of November, 2023, following primary elections that occurred on the 2nd of May. Joe Hogsett, the incumbent Democrat, successfully ran for a third term. Both Hogsett and his Republican challenger, Jefferson Shreve, made it through to the final round of the election.
    
\end{quote}
\section{Current Events Existing Knowledge Examples}\label{appendix:current_events_analysis}

To give a better understanding of how a model might be able to answer questions about new information, with better than random success, we present three possible scenarios as examples. These scenarios show how models with stronger reasoning skills can infer the correct answer even for unseen information.
\\\\
The first scenario involves questions about previously unseen information, where basic reasoning abilities allow a model to make an educated guess.

\begin{quote}
    \textbf{\underline{Question:}} What was a key issue that led to the 2023 United Auto Workers strike?
    \\\\
    \textbf{\underline{Answers:}} 
    \begin{enumerate}
        \item Dissatisfaction with the quality of cafeteria food.
        \item Disagreements over employee dress codes.
        \item Discontent with stagnant wages and tiered employment systems.
        \item Debates over the color scheme of the factories.
    \end{enumerate}
\end{quote}

In this case it is easy to guess that the third option is the most likely, even without knowledge of this specific strike.
\\\\
A second scenario involves questions where prior knowledge about a topic may aid a model in answering. 

\begin{quote}
    \textbf{\underline{Question:}} What environmental concern was raised by some scientists as a result of the 2023 Hawaii wildfires?
    \\\\
    \textbf{\underline{Answers:}}
    \begin{enumerate}
        \item Rising temperatures.
        \item Melting ice caps.
        \item Charred soils running off into the shoreline.
        \item Increased air pollution.
    \end{enumerate}
\end{quote}
In this case, knowing the geography of Hawaii, as well as immediate effects of wildfires, enables a model to give the first two options a lower likelihood. This process of elimination increases the probability of choosing one of the remaining options (the third option is the correct answer).
\\\\
A third scenario arises due to the automatic question generation process, some questions strongly rely on pre-existing knowledge.

\begin{quote}
    \textbf{\underline{Question:}} What event in 2021 was compared to the September 2023 New York floods?
    \\\\
    \textbf{\underline{Answers:}} 
    \begin{enumerate}
        \item Hurricane Katrina.
        \item Hurricane Ida.
        \item Hurricane Sandy.
        \item Hurricane Harvey.
    \end{enumerate}
\end{quote}

Since only one of these events occurred in 2021 (Hurricane Ida), and all the models tested have been exposed to events from 2021 during pre-training, this question can potentially be answered without using additional current information.
\\\\
Finally, to demonstrate why it is reasonable to assume that models cannot generally answer questions about new information, with better than random success, look at the following example:

\begin{quote}
    \textbf{\underline{Question:}} How did Matthew Belk, a National Weather Service meteorologist, describe the September 2023 northeastern U.S. floods?
    \\\\
    \textbf{\underline{Answers:}} 
    \begin{enumerate}
        \item 50-year event.
        \item 100-year event.
        \item 200-year event.
        \item 500-year event.
    \end{enumerate}
\end{quote}

Even with some knowledge about floods and their statistical properties, it would be very difficult to guess that this specific meteorologist would call the flood a `200-year event'. This is especially true if the model was not exposed to information about the details of the flood. 

\end{document}